\pdfoutput=1

\documentclass[11pt]{article}

\usepackage[]{EMNLP2023}

\usepackage{times}
\usepackage{latexsym}

\usepackage[T2A,T1]{fontenc}

\usepackage[utf8]{inputenc}

\usepackage{CJKutf8}

\usepackage[russian,english]{babel}

\usepackage{microtype}

\usepackage{inconsolata}

\usepackage{amsmath,amssymb}
\usepackage{arydshln}
\usepackage{array,multirow}
\usepackage{booktabs, hhline}
\usepackage{enumitem}
\usepackage{graphicx}
\usepackage{makecell}
\usepackage{subcaption}

\graphicspath{{figures/}}

\DeclareMathOperator*{\argmax}{arg\,max}

\title{MetricX-24: The Google Submission to the WMT 2024\\Metrics Shared Task}

\author{Juraj Juraska, Daniel Deutsch, Mara Finkelstein \and {\bf Markus Freitag} \\
  Google \\
  \texttt{\{jjuraska,dandeutsch,marafin,freitag\}@google.com}
}

\begin{document}

\maketitle

\begin{abstract}
    In this paper, we present the MetricX-24 submissions to the WMT24 Metrics Shared Task and provide details on the improvements we made over the previous version of MetricX. Our primary submission is a hybrid reference-based/-free metric, which can score a translation irrespective of whether it is given the source segment, the reference, or both. The metric is trained on previous WMT data in a two-stage fashion, first on the DA ratings only, then on a mixture of MQM and DA ratings. The training set in both stages is augmented with synthetic examples that we created to make the metric more robust to several common failure modes, such as fluent but unrelated translation, or undertranslation. We demonstrate the benefits of the individual modifications via an ablation study, and show a significant performance increase over MetricX-23 on the WMT23 MQM ratings, as well as our new synthetic challenge set.\footnote{
        Our code and models can be found at \url{https://github.com/google-research/metricx}.
    }
\end{abstract}
\section{Introduction}
\label{sec:introduction}

Automatic evaluation metrics are critical to the development of machine translation (MT) systems. In recent years, the landscape of MT evaluation has changed dramatically since the use of lexical metrics, like BLEU~\citep{papineni-etal-2002-bleu} and ChrF~\citep{popovic-2015-chrf}, that compared the tokens or characters of the candidate translation to a reference translation to predict a scalar score that represents the quality of the translation. Evaluation metrics based on neural networks opened up the door for more experimentation, and metrics now vary based on what type of output they produce, what they require as input for prediction, and whether they use a dedicated evaluation model or a general-purpose large language model.

This paper provides details on MetricX-24, the successor to MetricX-23. MetricX is a learned regression-based metric trained to predict a floating point score representing the quality of a translation. This year, we made four submissions to the WMT24 Metrics Shared Task, all based on the mT5 language model~\citep{xue-etal-2021-mt5}, which is further fine-tuned on direct assessment (DA) ratings, MQM ratings~\citep{lommel2014multidimensional,freitag-etal-2021-experts}, and newly constructed synthetic data. The primary submission, denoted MetricX-24-Hybrid, is a hybrid reference-based/-free metric, which can score a translation irrespective of whether it is given the source segment, the reference, or both. The same model is thus the primary submission for both the reference-based evaluation and the quality estimation (QE) task, having predicted the scores once with and once without the references provided in the input. Our contrasting submissions, MetricX-24(-QE), are standalone reference-based/QE models, trained only for their specific task.

The key takeaways from our experiments, detailed in this report, include:
\begin{enumerate}[itemsep=-0.25em]
    \item Learned metrics cannot reliably detect undertranslation, duplication, missing punctuation, and fluent but unrelated translation;
    \item Adding a relatively small amount of synthetic data to the training set can boost the metric's performance, especially on lower-quality translations with the above issues;
    \item It is possible to effectively train a metric on a mixture of MQM and DA ratings, thus maintaining high performance on a larger set of language pairs;
    \item Training a metric in the hybrid input mode, i.e., with and without the reference included in the input, allows it to learn to rely less on the reference when it is of poor quality.
\end{enumerate}

\section{Data}
\label{sec:data}

Developing MetricX-24, we relied solely on publicly available data from the WMT Metrics shared tasks between 2015 and 2023. The translation ratings from these years come in two different flavors: (1) direct assessment (DA) scores on a scale from 0 to 100, collected in general from non-expert raters, and (2) MQM scores~\citep{lommel2014multidimensional,freitag-etal-2021-experts} on a scale from 0 to 25 (with 0 being the best), which are grounded on error spans and their corresponding severity levels, annotated by professional raters. MQM ratings have been collected as part of the WMT campaign only since 2020 and, because the annotations are considerably more time-consuming and expensive to obtain, they are only available for a few language pairs. The DA scores, on the other hand, offer a broader language coverage of nearly 50 language pairs, but the raw ratings are noisy (due to different rating strategies) and generally of lower quality. Therefore, it is often beneficial to $z$-normalize DA ratings per rater before training models on them, so as to make the ratings more comparable across different annotators. In contrast, models do not benefit from MQM scores being $z$-normalized because the scores come from a rather small group of annotators and they adhere to a rubric.

In the rest of this section, we provide details on which data we use for training and evaluation, as well as how the different datasets are preprocessed. Furthermore, we describe new synthetic data we created from the WMT datasets, with the goal of addressing some of MetricX's known failure modes.


\subsection{Training Data}

\paragraph{DA.} We utilize most of the DA data from the 2015--2022 period for training, with the following exceptions. As we observed during the development of the previous version of MetricX~\citep{juraska-etal-2023-metricx}, the into-English portion of the WMT21 DA ratings drags the model performance down. We confirmed this observation again this year and excluded these language pairs from the training data. With the gradually declining quality of DA ratings collected for WMT using the MTurk platform, we also exclude all into-English language pairs from WMT22.\footnote{One exception is zh-en, for which DA ratings were collected in two different ways, including using the same method and framework as the out-of-English language pairs~\citep{kocmi-etal-2022-findings}.} Additionally, we exclude the en-zh language pair from WMT22, as we use the equivalent slice of data, but with MQM ratings, for evaluation. We use $z$-normalized ratings when training models on DA data only, but raw ratings when training on a mixture of MQM and DA data.

\paragraph{MQM.} Besides the DA ratings, we also take advantage of the higher-quality MQM ratings from the years up to 2022 for training. These include four language pairs: en-de, en-ru, en-zh and zh-en.\footnote{The en-zh MQM ratings, available at \url{https://github.com/google/wmt-mqm-human-evaluation}, were collected post-WMT22.} We only use the conversation, e-commerce and social domains from WMT22 en-zh for training. In our experiments with different subsets of MQM ratings, we observed a consistent boost in performance with the 2020 data excluded, hence, our final models are only trained on MQM ratings from 2021 and 2022. We always train models on raw MQM ratings, i.e., using the 0--25 scale.

\subsection{Evaluation Data}
\label{sec:evaluation_data}

\paragraph{MQM.} Our primary evaluation set consists of the WMT23 MQM ratings, which includes three language pairs: en-de, he-en and zh-en. Since the zh-en language pair is known to have low-quality references~\citep{kocmi-etal-2023-findings}, we replace them with newly collected references. Note that this has no effect on the MQM ratings, as those were collected in a source-based fashion.
Additionally, given the fact that one of the official WMT24 test language pairs is ja-zh, we reserve the news domain subset of the WMT22 en-zh ratings for evaluation, allowing us to assess our models' performance on a language pair with Chinese as the target language.


\paragraph{DA.} We use the WMT23 DA ratings as a secondary evaluation set, taking advantage of its better language coverage (8 language pairs). Nevertheless, with DA ratings generally following a significantly different distribution than MQM ratings, a higher correlation of the metric scores with these DA ratings does not necessarily imply better performance. For example, fine-tuning a model on zh-en MQM ratings results in lower performance than fine-tuning it on DA ratings, according to the zh-en DA evaluation set (but not the MQM one). Therefore, we only consider the WMT23 DA evaluation set in experiments where we mix MQM and DA training data together.

\subsection{Synthetic Data}
\label{sec:synthetic_data}

After seeing the initial benefits from the simple synthetic data used for training MetricX-23, we decided to construct a more comprehensive collection of synthetic training examples. They cover additional, less trivial failure modes of MetricX, i.e., translation issues commonly unrecognized by the metric.
The DEMETR challenge set~\citep{karpinska-etal-2022-demetr}, which we relied on last year, does not cover several of the failure modes we created the new synthetic training examples for, hence we also constructed a set of test examples for each of them. Next, we describe how we designed both the training and the test synthetic datasets.

\begin{table*}
    \small
    \centering
    \begin{tabular}{l m{0.65\linewidth} c}
        \toprule
        \textbf{Failure mode}   & \textbf{Synthetic candidate translation description}    & \textbf{MQM score} \\
        \midrule
        Empty translation   & Empty string. & 25 \\
        \midrule
        Gibberish   & Text of a similar length as the reference, generated by sampling words from the vocabulary built from all references in the data with a matching target language.   & 25 \\
        \midrule
        \makecell[l]{Fluent but unrelated\\translation}  & Arbitrary reference from the dataset of a similar length and in the same language.   & 25 \\
        \midrule
        Undertranslation    & Candidate translation with an arbitrary sentence removed, if a multi-sentence segment, otherwise, candidate translation with 20--80\% words removed from the end.    & 5--25 \\
        \midrule
        Duplication & Candidate translation duplicated, with a space in between.    & 25 \\
        \midrule
        Missing punctuation & Reference translation with the end punctuation removed (11 punctuation symbols considered, such as period, question mark, closing parenthesis or quotation mark). & 1 \\
        \midrule
        \makecell[l]{Reference-matching\\translation}   & Reference translation itself (unlike the rest, these synthetic examples are meant to train the metric to predict a perfect score for translations matching the reference). & 0 \\
        \bottomrule
    \end{tabular}
    \vspace{-0.05in}
    \caption{Failure mode categories we prepared synthetic data for, along with brief descriptions of how we created the synthetic examples from the WMT data, and the MQM scores we label the training examples with.}
    \label{tab:synthetic_data_overview}
    \vspace{-0.1in}
\end{table*}

\subsubsection{Training Sets}

In order for the MetricX models to learn to identify certain types of bad translations that are not sufficiently (or at all) represented in the regular WMT training data, we generated synthetic examples that we augment the training data with. They were created by modifying examples from the DA datasets ranging from WMT15 to WMT22, comprising 49 language pairs. Table~\ref{tab:synthetic_data_overview} provides an overview of the various failure modes that we considered, including brief descriptions of how we prepared the synthetic data to address them. Additional details regarding the creation process can be found in Appendix~\ref{app_sec:synthetic_data_creation}.

\subsubsection{Test Set}

We constructed a new DEMETR-style test set based on the WMT23 DA dataset, with examples generated analogously to our synthetic training examples, as described in Table~\ref{tab:synthetic_data_overview}. Each synthetic example is paired with its original counterpart (although using the reference instead of the candidate translation whenever the synthetic translation was created from the reference), which allows for a metric to be evaluated on how frequently it ranks the pairs correctly.

\section{Metric Descriptions}
\label{sec:metric_description}

The MetricX-24 submissions to the WMT24 Metrics Shared Task build on top of the successful MetricX-23~\citep{juraska-etal-2023-metricx,kocmi-etal-2023-findings}, with several major improvements. We start this section by summarizing the aspects this year's submissions have in common with MetricX-23, then provide an overview of the modifications, and finally describe the differences between the individual submissions.

\subsection{MetricX Model}

MetricX is a learned metric, powered by a regression model trained to predict a floating point number that represents the quality of a given translation. The reference-based variant takes the candidate translation (hypothesis) and reference segments as input, and concatenates them, along with corresponding prefixes (``candidate:'' and ``reference:'', respectively). In contrast to the previous versions, MetricX-24 also prepends the source segment (along with the prefix ``source:'') to the input, offering the model additional context to make a better prediction in the reference-based setting, which may be beneficial especially in cases where the reference is inadequate. The model then encodes this combined input and uses it to predict the translation quality score. The QE variant works in an analogous way, but taking only the source segment and the hypothesis as the input.

With MetricX-24, we continue to rely on mT5~\citep{xue-etal-2021-mt5} as the pretrained language model that we fine-tune on translation evaluation data. We refer the reader to~\citet{juraska-etal-2023-metricx} for details on how we adapted this encoder-decoder model to the regression task. Similar to MetricX-23, we fine-tune the model in two stages: first on DA ratings ($z$-normalized, aggregated per segment, negated, and finally clipped to the $[-1.0, 1.0]$ range) and then further on raw MQM ratings. As a result, the metric produces scores in the [0, 25] range. The model is trained with a mean squared error (MSE) loss function. Further implementation details can be found in~\S\ref{sec:setup}.

\subsection{Design Improvements}

We achieve some initial improvement in performance by simply including the WMT22 data in the training set -- both the DA and the MQM ratings, which we previously used as the evaluation set when developing MetricX-23. The additional MQM ratings (including en-ru, a language pair not present in the older MQM data) are especially valuable, considering the scarcity of MQM data. Besides that, we introduce three major modifications to the training procedure and data in order to further improve MetricX's performance, described throughout the rest of this section.

\subsubsection{Training With Synthetic Data}
\label{sec:training_with_synthetic_data}

Although we used synthetic training data alongside the DA and MQM ratings already for training MetricX-23, the synthetic examples covered only the two trivial cases of empty and reference-matching translations. As described in~\S\ref{sec:synthetic_data}, we prepared a significantly more comprehensive synthetic training set for MetricX-24, which we combine with the DA and MQM data in both fine-tuning stages. We experimented with various ratios, and settled on $1$:$100$ for each synthetic example category in the first stage and $1$:$5000$ in the second stage. We evaluate the effects of adding the synthetic training data by measuring accuracy and average score differences on the synthetic test set, also described in~\S\ref{sec:synthetic_data}.

\subsubsection{Mixing DA and MQM Data}
\label{sec:mixing_da_and_mqm_data}

Next, we attempt to address the inevitable decline in MetricX performance on other languages after fine-tuning the model on MQM data, which only covers a few language pairs. The performance, as measured by the WMT23 DA evaluation set with 8 language pairs, quickly declines after starting to fine-tune on MQM ratings. While it is expected that the change in the general score distribution -- caused by the switch from DA to MQM ratings -- results in the Pearson correlations with the ground-truth scores dropping, we believe the model should be able to retain its system- and segment-level pairwise accuracy from the first stage of fine-tuning on DA data. Moreover, we observe a significant drop in system-level performance on the zh-en language pair of the MQM evaluation set, despite zh-en being present in the MQM training data.

In order to remedy these behaviors, we mix in a smaller proportion of DA ratings in the second-stage fine-tuning. That way the model is trained primarily on MQM ratings, but has a continued exposure to the additional 40+ language pairs from the first stage of fine-tuning. We experimented with different combinations of DA and MQM rating formats (e.g., raw vs.~$z$-normalized, transformed to the MQM scale or not, etc.), and the one yielding the best results was raw MQM ratings combined with raw DA ratings linearly transformed to the MQM scale of $[0, 25]$. Finally, we determined that a DA:MQM ratio of $1$:$4$ works well for boosting the performance on the DA evaluation set back to the levels from the first stage of fine-tuning, without a significant negative impact on the model's performance on the MQM evaluation set.\footnote{After a more extensive post-submission experimentation, we determined the optimal ratio to be $1$:$10$.}


\subsubsection{Hybrid Input Mode}

The third major modification we make to the training procedure when developing MetricX-24, is mixing training examples in three different formats: (1)~source + hypothesis, (2)~hypothesis + reference, and (3)~source + hypothesis + reference. This allows the model to operate in both a QE and a reference-based mode (and the latter either with or without the source included). But perhaps more importantly, it gives the model an opportunity to learn how much weight to put on the source and the reference in different scenarios, or possibly to completely ignore the reference when it is of low quality. Such a hybrid model is then evaluated as a QE model by only passing it the source segment and the hypothesis as input, and as a reference-based model by additionally passing it the reference.

\subsection{MetricX-24 Variants}

There are four variants of MetricX-24 that we submitted to the WMT24 Metrics Shared Task:
\begin{itemize}[itemsep=-0.25em]
    \item MetricX-24-Hybrid (primary)
    \item MetricX-24-Hybrid-QE (primary)
    \item MetricX-24
    \item MetricX-24-QE
\end{itemize}
Our primary reference-based and QE submissions are actually the same hybrid model, with the scores predicted with and without the references provided as part of the input. The secondary submissions are the standalone reference-based and QE counterparts of the hybrid model, i.e., only trained on examples with the references (as well as the source segments) included and on examples with the references omitted, respectively. Other than that, all of the submission models are identical in terms of training data mixtures, as described in~\S\ref{sec:training_with_synthetic_data} and~\S\ref{sec:mixing_da_and_mqm_data}, as well as training hyperparameters.

\section{Experimental Setup}
\label{sec:setup}

\subsection{Meta-Evaluation}

As mentioned in~\S\ref{sec:evaluation_data}, our primary evaluation set consists of the MQM ratings from WMT23, as well as the news domain subset of the en-zh language pair from WMT22. Considering there is no into-English language pair among the official test sets this year, we focus primarily on en-de and en-zh when evaluating our models, but also keeping zh-en (the dataset with alternate references) in the mix, in order to ensure that we do not overfit the models to out-of-English language pairs. To evaluate our models, we calculate the agreements between their predicted scores and the human judgments of translation quality using the four different correlations from the WMT23 Metrics Shared Task~\citep{freitag-etal-2023-results}, detailed in Appendix~\ref{app_sec:meta_eval_details}.

\subsection{Checkpoint Selection}

In both the first and the second stage of fine-tuning, we pick the best checkpoint $c_\textsubscript{best}$ based on the following linear combination of segment- and system-level pairwise accuracy:
\[ \argmax_{c} \ \frac{3}{4} \sum_{l} \text{acc}_{l}^\textsuperscript{seg}(c) + \frac{1}{4} \sum_{\text{l}} \text{acc}_{l}^\textsuperscript{sys}(c) \ , \]
where $l$ $\in$ \{en-de, en-zh, zh-en\}, and $\text{acc}_{l}^\textsuperscript{seg}(c)$ and $\text{acc}_{l}^\textsuperscript{sys}(c)$ are the segment- and the system-level pairwise accuracy calculated for checkpoint $c$ on the language pair $l$ of the evaluation set. We downweight the system-level component to account for its greater variance and to thus avoid a checkpoint being picked due to a rare spike in system-level accuracy if segment-level accuracy is low.

\subsection{Implementation Details}

MetricX-24, similar to its predecessor, is implemented with TensorFlow~\citep{tensorflow2015-whitepaper} and the T5X library~\citep{roberts2023scaling}. All of the metric variants are based on mT5-XXL with 13B parameters. We defer further implementation details to Appendix~\ref{app_sec:implementation_details}. We are publicly releasing our submissions, converted from TensorFlow to PyTorch~\citep{paszke2019pytorch} checkpoints.\footnote{\url{https://github.com/google-research/metricx}}

\section{Results and Discussion}
\label{sec:results}

Here we present the results of our experiments, focusing solely on fully trained models (i.e., those that went through both stages of fine-tuning) and modifications in the second stage. Since the ablation studies performed with reference-based and QE models show similar trends, we discuss the reference-based experiments in depth in this section, and provide the QE results in Appendix~\ref{app_sec:qe_results}. Due to limited resource availability, we were only able to run each experiment with one random seed.

\subsection{Training With Synthetic Data}

We start by examining the benefits of including synthetic training examples, as described in~\ref{sec:synthetic_data}. In Table~\ref{tab:synthetic_results}, the bottom four rows -- corresponding to the hybrid model -- demonstrate the effects of progressively adding DA data only, synthetic data only, and finally both, to the training set in the second stage of fine-tuning.\footnote{The models that did not include synthetic training data in the second stage, did not use it in the first stage either.} We ended up not using the duplication synthetic training set, as we observed that the models learn to correctly identify such cases even without it.

\begin{table*}
    \small
    \centering
    \begin{tabular}{>{\centering\arraybackslash} m{0.08\linewidth} >{\centering\arraybackslash} m{0.05\linewidth} >{\centering\arraybackslash} m{0.075\linewidth} >{\centering\arraybackslash} m{0.075\linewidth} >{\centering\arraybackslash} m{0.076\linewidth} >{\centering\arraybackslash} m{0.075\linewidth} >{\centering\arraybackslash} m{0.076\linewidth} >{\centering\arraybackslash} m{0.075\linewidth} >{\centering\arraybackslash} m{0.076\linewidth} >{\centering\arraybackslash} m{0.075\linewidth}}
        \toprule
        \textbf{MetricX variant} & \textbf{+DA} & \textbf{+Synth} & \textbf{Empty transl.} & \textbf{Gib-berish} & \textbf{Unre-lated} & \textbf{Under-transl.} & \textbf{Dupli-cation} & \textbf{Missing punct.} & \textbf{Ref-match} \\
        \midrule
        23 & -- & $\sim$ & \textbf{100.00} & \textbf{100.00} & 88.14 & 57.75 & 38.14 & 66.01 & \textbf{94.00} \\
        \midrule
        24 & \checkmark & \checkmark & 99.29 & 99.86 & \textbf{99.29} & \textbf{98.75} & 99.14 & 83.01 & 78.14 \\
        \midrule
        \multirow{4}{*}{\rotatebox[origin=c]{90}{24-Hybrid}} & -- & -- & 51.43 & 99.86 & 81.00 & 68.75 & 87.57 & 83.66 & 76.00 \\
        & \checkmark & -- & 53.57 & 99.71 & 92.14 & 82.25 & \textbf{99.57} & \textbf{85.62} & 72.86 \\
        & -- & \checkmark & 94.14 & 99.71 & 99.14 & 96.25 & 94.43 & 84.97 & 79.86 \\
        & \checkmark & \checkmark & 97.29 & 99.71 & 98.71 & 96.25 & 99.43 & 82.35 & 75.14 \\
        \bottomrule
    \end{tabular}
    \vspace{-0.05in}
    \caption{Accuracy of reference-based MetricX variants in all 7 categories of our synthetic test set. ``23'' is the baseline, the last row of ``24-Hybrid'' corresponds to our primary submission, and ``24'' is our secondary submission.}
    \label{tab:synthetic_results}
\end{table*}

\begin{table*}
    \small
    \centering
    \begin{tabular}{>{\centering\arraybackslash} m{0.08\linewidth} >{\centering\arraybackslash} m{0.05\linewidth} >{\centering\arraybackslash} m{0.075\linewidth} >{\centering\arraybackslash} m{0.066\linewidth} >{\centering\arraybackslash} m{0.06\linewidth} >{\centering\arraybackslash} m{0.067\linewidth} >{\centering\arraybackslash} m{0.06\linewidth} >{\centering\arraybackslash} m{0.066\linewidth} >{\centering\arraybackslash} m{0.06\linewidth} >{\centering\arraybackslash} m{0.067\linewidth} >{\centering\arraybackslash} m{0.06\linewidth}}
        \toprule
        \multirow{2}{*}{\textbf{\makecell{MetricX\\variant}}} & \multirow{2}{*}{\textbf{+DA}} & \multirow{2}{*}{\textbf{+Synth}} & \multicolumn{4}{c}{\textbf{Segment-level pairwise accuracy}} & \multicolumn{4}{c}{\textbf{System-level pairwise accuracy}} \\
        & & & \textbf{en-de} & \textbf{zh-en} & \textbf{zh-en\textsuperscript{\textdagger}} & \textbf{en-zh} & \textbf{en-de} & \textbf{zh-en} & \textbf{zh-en\textsuperscript{\textdagger}} & \textbf{en-zh} \\
        \midrule
        23 & -- & $\sim$ & 60.20 & 53.12 & 54.06 & 55.73 & 90.91 & 89.52 & 86.67 & 74.36 \\
        \midrule
        24 & \checkmark & \checkmark & 60.71 & 54.50 & 55.78 & 56.16 & 96.97 & \textbf{92.38} & \textbf{95.00} & \textbf{88.46} \\
        \midrule
        \multirow{4}{*}{\rotatebox[origin=c]{90}{24-Hybrid}} & -- & -- & 61.17 & 54.63 & 55.52 & 57.43 & \textbf{100.00} & 89.52 & 91.67 & 85.90 \\
        & \checkmark & -- & 60.75 & 54.89 & 55.58 & 57.65 & 98.48 & \textbf{92.38} & 92.50 & 84.62 \\
        & -- & \checkmark & \textbf{61.75} & 54.38 & 55.43 & \textbf{57.73} & 98.48 & 90.48 & 91.67 & \textbf{88.46} \\
        & \checkmark & \checkmark & 61.11 & \textbf{55.00} & \textbf{55.82} & 57.02 & 98.48 & \textbf{92.38} & 94.17 & 85.90 \\
        \bottomrule
    \end{tabular}
    \vspace{-0.05in}
    \caption{Meta-evaluation scores of reference-based MetricX variants on the WMT23 MQM evaluation set. ``23'' is the baseline, the last row of ``24-Hybrid'' corresponds to our primary submission, and ``24'' is our secondary submission. \textsuperscript{\textdagger}Alternate references.}
    \label{tab:meta_eval_results}
\end{table*}

The first thing to notice is that mixing in DA ratings actually improves the metric's performance on the synthetic test set over fine-tuning on MQM ratings alone, especially in the unrelated, undertranslation and duplication failure modes. Adding synthetic data instead is, however, significantly more effective in general, boosting the accuracy to the $94$--$100$\% range in most categories. Finally, augmenting the training set with both the DA and the synthetic data results in an overall similar performance as with the synthetic data only.

Missing punctuation is one of two categories in which our metrics score not so close to perfect. In fact, the synthetic training examples appear not to be helpful in improving the performance at all. Our hypothesis is that using references to create this category of synthetic examples results in a significant proportion of misleading examples because we assume references to be perfect, but that is not always the case. That, combined with the fact that the removal of the punctuation symbol from the end of the segment warrants just a minor score change, means that some of the synthetic examples might have an unreasonably high ground-truth score associated with them, thus giving the model the opposite signal to what is desired.

The reference-matching translation synthetic training set appears not to be effective either, however, its benefits are somewhat concealed by the fact that mixing in DA data drags the performance in this category down. With the non-hybrid model, we observed a significantly bigger drop with DA data included (77\% $\rightarrow$ 64\%) and a greater increase with synthetic data included instead (77\% $\rightarrow$ 83\%). Granted, that is still far from perfect, however, expecting a 100\% accuracy in this category equates to expecting that the candidate translation is never better than the reference, which, as we pointed out earlier, is not always true when judging the translation quality based on the source segment.

Overall, thanks to the new synthetic training data, MetricX-24 (hybrid or not) is clearly more robust to the failure modes than MetricX-23 (see first row in the table), with the reference-matching translations being an exception. That might have to do with the absence of WMT22 data in the training set of MetricX-23, or the only synthetic examples present therein being those of empty and reference-matching translations.

\subsection{Mixing DA and MQM Data}

We already discussed the effects of adding DA data to the training set in the second stage of fine-tuning in terms of the synthetic test set performance; let us now have a look at the correlations with human MQM scores. Comparing the first two rows of the ``24-Hybrid'' section in Table~\ref{tab:meta_eval_results}, we see that there are just relatively minor changes in either direction across all language pairs, with score differences within the expected variance between runs.

What the table does not show, however, is the huge jump in all correlations across all language pairs on the WMT23 DA evaluation set, typically back to the levels from the first stage of fine-tuning on DA data only, or above. Segment- and system-level pairwise accuracy increases by up to 2 and 5 points, respectively, and Pearson's $r$ sees improvements of up to 10 points. These are valuable gains, considering we achieved them without sacrificing the performance on the MQM evaluation set. An overview of the results and a more detailed analysis on the DA evaluation set can be found in Appendix~\ref{app_sec:mixing_da_and_mqm_data_results}.

\subsection{Hybrid Input Mode}

To wrap up the evaluation, we discuss the performance difference between MetricX-24 and MetricX-24-Hybrid (rows 2 and 6 in Table~\ref{tab:meta_eval_results}). At the system level, the hybrid variant lags slightly behind in zh-en and en-zh, but it makes up for it by outperforming the non-hybrid across the board at the segment level. Notably, the hybrid metric achieves an almost 1\% higher segment-level accuracy on en-zh, and the 0.5\% boost on zh-en (with original references) may be evidence for the hybrid model handling examples with poor-quality references better, especially considering the accuracy difference on the zh-en set with alternate references is only 0.04\%. The other performance differences between the two models are largely insignificant.

Finally, comparing our primary submission with MetricX-23 (row 1 in the table), we can see consistent gains of 1--2 points in segment-level accuracy, and substantially bigger gains at the system level, with the accuracy on en-zh improving by a whopping 11.5 points. We conclude that this a significant improvement over our last year's submission, ranked overall second in the WMT23 Metrics Shared Task.

\section{Related Work}
\label{sec:related_work}

Traditionally, evaluation metrics predict a scalar quality score for the translation.
This type of metric includes BLEU, ChrF, MetricX~\citep{juraska-etal-2023-metricx}, BLEURT~\citep{sellam-etal-2020-bleurt,pu-etal-2021-learning}, COMET~\citep{rei-etal-2020-comet,rei-etal-2022-comet}, COMETKiwi~\citep{rei-etal-2022-cometkiwi}, Prism \citep{thompson-post-2020-automatic}, and more.
While these metrics have historically been the dominant category of metric, newly proposed methods provide structured \citep{perrella-etal-2022-matese,fernandes-etal-2023-devil,kocmi-federmann-2023-large,guerreiro2023xcomet} or natural language explanations \citep{xu-etal-2023-instructscore} for the predicted scores.

Then, evaluation metrics are considered to be reference-based or reference-free (also known as ``quality estimation'') depending on whether or not they require a reference to evaluate a translation.
Metric developers usually train separate models for each type of metric (e.g., COMET and COMETKiwi, or MetricX-23 and MetricX-23-QE), but some opt for combining both tasks into a single model \citep{wan-etal-2022-alibaba,guerreiro2023xcomet}, which is the approach we took in this work with our hybrid model.

Finally, while most metrics like MetricX-24 use a dedicated model for scoring translations, some recent works have begun to leverage general-purpose large language models instead \citep{fernandes-etal-2023-devil,kocmi-federmann-2023-large,xu-etal-2023-instructscore,leiter-etal-2023-eval4nlp,leiter2024prexme}.
While LLM-based metrics have achieved strong system-level performance, using a learned dedicated model was the best approach at the segment-level in last year's Metrics Shared Task \citep{freitag-etal-2023-results}.

\section{Conclusion}
\label{sec:conclusion}

We presented in detail our approach to training MetricX-24, a regression-based MT evaluation metric. We submitted four versions of MetricX-24 to the WMT24 Metrics Shared Task, including a reference-based and a QE variant, as well as a new hybrid variant evaluated with and without the references. By evaluating on the WMT23 MQM dataset, we showed all of them to significantly outperform our last year's submission, MetricX-23. In addition, we made MetricX-24 more robust to various types of bad translations, which do not frequently occur in the WMT data, such as undertranslation, or fluent but unrelated translation. Finally, by combining DA and MQM ratings together in the final stage of fine-tuning, we were able to dramatically increase the performance on the WMT23 DA dataset covering 8 language pairs, while maintaining the high correlations with the MQM ratings at the same time.

\bibliography{custom}
\bibliographystyle{acl_natbib}

\clearpage

\appendix
\section{Synthetic Data Creation}
\label{app_sec:synthetic_data_creation}

We sample 500 examples from each language pair, whose candidate translations (hypotheses) we then manipulate in different ways to create the synthetic examples for each failure mode category. The missing punctuation category is an exception, with a stratified sample across the 11 end-punctuation symbols, rather than language pairs, and 250 examples each.

In general, the synthetic examples have the candidate translation manipulated, turning it into a worse, or an outright bad, translation. One exception is the reference-matching category, whose purpose is to actually teach the metric to score translations that match the reference highly, which it does not learn to do reliably when only trained on the WMT data. Table~\ref{tab:synthetic_data_examples} shows a few concrete examples from the synthetic training set.

\begin{table*}
    \centering
    \begin{tabular}{l m{0.9\linewidth}}
        \toprule
        \multicolumn{2}{l}{\textbf{Gibberish (zh-en example)}} \\
        \multicolumn{2}{l}{\emph{Created from: corpus hypothesis vocabulary}} \\
        src & \begin{CJK*}{UTF8}{gbsn}我希望你们能准时，不是想要你们的优惠券！！\end{CJK*} \\
        hyp & filter two that to also in allegations train 800 city, continuous the \\
        ref & I hope you can be on time, and it’s not that I want your coupons! ! \\
        label & 25 \\
        \midrule
        \multicolumn{2}{l}{\textbf{Fluent but unrelated translation (de-en example)}} \\
        \multicolumn{2}{l}{\emph{Created from: corpus references}} \\
        src & Damit können doppelt so viele Studierende ausgebildet werden wie bisher. \\
        hyp & She booked a return flight and went home the next day. \\
        ref & In that way, twice as many students can be educated as before. \\
        label & 25 \\
        \midrule
        \multicolumn{2}{l}{\textbf{Undertranslation (cs-en example)}} \\
        \multicolumn{2}{l}{\emph{Created from: hypothesis}} \\
        src & Dlouhodobě napjaté vztahy mezi oběma zeměmi se vyostřily v roce 2018 poté, co Washington odstoupil od jaderné dohody z roku 2015 mezi Íránem a světovými mocnostmi a zavedl vůči Íránu sankce, které mají tvrdý dopad na jeho ekonomiku. \\
        hyp & Long-tense relations between the two countries sharpened in 2018 after Washington withdrew from the 2015 nuclear deal between Iran and world powers and imposed sanctions. \\
        ref & Long-term tense relations between both countries escalated in 2018 after that Washington withdrew from the nuclear deal closed in 2015 between Iran and the world powers and imposed sanctions against Iran, which have had hard impacts on its economy. \\
        label & 12.75 \\
        \midrule
        \multicolumn{2}{l}{\textbf{Duplication (fi-en example)}} \\
        \multicolumn{2}{l}{\emph{Created from: hypothesis}} \\
        src & Ensi vuoden vaje on yli 2,4 prosenttia kansantuotteesta. \\
        hyp & Next year's deficit will be over 2.4 per cent of national product. Next year's deficit will be over 2.4 per cent of national product. \\
        ref & Next year's deficit is over 2.4 per cent of GDP. \\
        label & 15 \\
        \midrule
        \multicolumn{2}{l}{\textbf{Missing punctuation (ru-en example)}} \\
        \multicolumn{2}{l}{\emph{Created from: reference}} \\
        src & \begin{otherlanguage*}{russian}Последний альбом Ace вышел в 2016 году.\end{otherlanguage*} \\
        hyp & Their last album, “Ace”, came out in 2016 \\
        ref & Their last album, “Ace”, came out in 2016. \\
        label & 1 \\
        \midrule
        \multicolumn{2}{l}{\textbf{Reference-matching translation (ja-en example)}} \\
        \multicolumn{2}{l}{\emph{Created from: reference}} \\
        src & \begin{CJK*}{UTF8}{gbsn}グレタさんは、27日の金曜日にも行うことを呼びかけていた。\end{CJK*} \\
        hyp & Now, Greta is calling for further strikes to be held on Friday the 27th. \\
        ref & Now, Greta is calling for further strikes to be held on Friday the 27th. \\
        label & 0 \\
        \bottomrule
    \end{tabular}
    \vspace{-0.05in}
    \caption{Synthetic examples for the different failure mode categories (except for the trivial empty translation case), along with the MQM scores we label the training examples with. Each category also has an indication of how the hypothesis was created/generated in order to produce a synthetic example (e.g., by modifying the original hypothesis or reference).}
    \label{tab:synthetic_data_examples}
    \vspace{-0.1in}
\end{table*}

\section{Meta-Evaluation Details}
\label{app_sec:meta_eval_details}

\paragraph{System-Level.} At the system level, we measure pairwise ranking accuracy~\citep{kocmi-etal-2021-ship}, as well as Pearson's~$r$. Pairwise accuracy assesses how well a metric ranks MT systems by calculating the proportion of all possible pairs of systems that are ranked the same by the metric and human scores. Pearson's~$r$, on the other hand, captures how strong the linear relationship is between the metric and human scores for MT systems. We obtain the system-level scores (both metric and human) as the mean segment-level score for each system.

\paragraph{Segment-Level.} At the segment level, we use the group-by-item pairwise accuracy with tie calibration, as described by~\citet{deutsch-etal-2023-ties}, and the no-grouping Pearson's~$r$. The pairwise accuracy calculates the proportion of all possible pairs of translations for the same source segment that are ranked the same by the metric and human, then averages the accuracies over all input segments. At the same time, it rewards correct tie predictions by introducing ties for any two translations with a metric score difference below an automatically determined threshold. The no-grouping Pearson's~$r$ quantifies the linear relationship between the metric and human scores across all translations from every system and document.

\section{Implementation Details}
\label{app_sec:implementation_details}

Having increased the maximum segment length from 256 to 512 SPM tokens, and including up to three segments (source, hypothesis and reference) in the model's input, each training run requires 256 TPUs. Using a batch size of 256, we train our models for 16K steps in the first stage, using a learning rate of 0.001 with an inverse square root decay after the first 2K steps. We then fine-tune the best checkpoint for another 8K steps in the second stage, lowering the learning rate to 0.0002 and decaying it after 1K steps. The models are trained using the Adafactor optimizer~\citep{shazeer2018adafactor}.

\section{Additional Results}

\subsection{Mixing DA and MQM Data}
\label{app_sec:mixing_da_and_mqm_data_results}

\begin{table*}
    \small
    \centering
    \begin{tabular}{>{\centering\arraybackslash} m{0.08\linewidth} >{\centering\arraybackslash} m{0.05\linewidth} >{\centering\arraybackslash} m{0.075\linewidth} >{\centering\arraybackslash} m{0.066\linewidth} >{\centering\arraybackslash} m{0.06\linewidth} >{\centering\arraybackslash} m{0.067\linewidth} >{\centering\arraybackslash} m{0.06\linewidth} >{\centering\arraybackslash} m{0.066\linewidth} >{\centering\arraybackslash} m{0.06\linewidth} >{\centering\arraybackslash} m{0.067\linewidth} >{\centering\arraybackslash} m{0.06\linewidth}}
        \toprule
        \multirow{2}{*}{\textbf{\makecell{MetricX\\variant}}} & \multirow{2}{*}{\textbf{+DA}} & \multirow{2}{*}{\textbf{+Synth}} & \multicolumn{4}{c}{\textbf{Segment-level pairwise accuracy}} & \multicolumn{4}{c}{\textbf{System-level pairwise accuracy}} \\
        & & & \textbf{en-de} & \textbf{zh-en} & \textbf{en-cs} & \textbf{de-en} & \textbf{en-de} & \textbf{zh-en} & \textbf{en-cs} & \textbf{de-en} \\
        \midrule
        DA only & N/A & N/A & 61.77 & 56.33 & 59.54 & 61.14 & \textbf{95.45} & 79.05 & \textbf{87.62} & 92.31 \\
        \midrule
        \multirow{4}{*}{\makecell{DA then\\MQM}} & -- & -- & 61.59 & 55.99 & 57.43 & 61.65 & 93.94 & 81.90 & 82.86 & 85.90 \\
        & \checkmark & -- & 61.88 & \textbf{56.67} & \textbf{60.16} & 62.29 & \textbf{95.45} & 80.95 & 86.67 & 88.46 \\
        & -- & \checkmark & \textbf{62.60} & 56.35 & 59.02 & 61.92 & \textbf{95.45} & \textbf{84.76} & 81.90 & \textbf{93.59} \\
        & \checkmark & \checkmark & 61.89 & 56.64 & 60.04 & \textbf{62.32} & 93.94 & 83.81 & 86.67 & \textbf{93.59} \\
        \bottomrule
    \end{tabular}
    \vspace{-0.05in}
    \caption{Meta-evaluation scores of reference-based MetricX variants on a subset of the language pairs of the WMT23 DA evaluation set. ``DA only'' is a model after just the first stage of fine-tuning (i.e., on DA data only), whereas the ``DA then MQM'' section contains models fine-tuned in full two stages. The last row thus corresponds to the ``24'' row in Tables~\ref{tab:synthetic_results} and~\ref{tab:meta_eval_results}, i.e., our secondary submission ``MetricX-24''.}
    \label{tab:meta_eval_results_da_eval_set}
\end{table*}

\begin{table*}
    \small
    \centering
    \begin{tabular}{>{\centering\arraybackslash} m{0.08\linewidth} >{\centering\arraybackslash} m{0.05\linewidth} >{\centering\arraybackslash} m{0.075\linewidth} >{\centering\arraybackslash} m{0.066\linewidth} >{\centering\arraybackslash} m{0.06\linewidth} >{\centering\arraybackslash} m{0.067\linewidth} >{\centering\arraybackslash} m{0.06\linewidth} >{\centering\arraybackslash} m{0.066\linewidth} >{\centering\arraybackslash} m{0.06\linewidth} >{\centering\arraybackslash} m{0.067\linewidth} >{\centering\arraybackslash} m{0.06\linewidth}}
        \toprule
        \multirow{2}{*}{\textbf{\makecell{MetricX\\variant}}} & \multirow{2}{*}{\textbf{+DA}} & \multirow{2}{*}{\textbf{+Synth}} & \multicolumn{4}{c}{\textbf{Segment-level Pearson's $r$}} & \multicolumn{4}{c}{\textbf{System-level Pearson's $r$}} \\
        & & & \textbf{en-de} & \textbf{zh-en} & \textbf{en-cs} & \textbf{de-en} & \textbf{en-de} & \textbf{zh-en} & \textbf{en-cs} & \textbf{de-en} \\
        \midrule
        DA only & N/A & N/A & \textbf{60.10} & \textbf{41.52} & \textbf{43.47} & 52.59 & 98.48 & \textbf{89.21} & \textbf{92.49} & 97.20 \\
        \midrule
        \multirow{4}{*}{\makecell{DA then\\MQM}} & -- & -- & 48.18 & 34.66 & 39.77 & 44.09 & 93.41 & 87.58 & 90.60 & 87.40 \\
        & \checkmark & -- & 53.67 & 36.29 & 43.11 & 52.79 & 93.67 & 87.75 & 90.40 & 91.53 \\
        & -- & \checkmark & 56.03 & 35.21 & 37.24 & 45.26 & \textbf{99.48} & 88.40 & 89.32 & 96.79 \\
        & \checkmark & \checkmark & 57.92 & 37.17 & 42.55 & \textbf{55.23} & 98.56 & 88.94 & 91.58 & \textbf{97.97} \\
        \bottomrule
    \end{tabular}
    \vspace{-0.05in}
    \caption{Same as Table~\ref{tab:meta_eval_results_da_eval_set}, but showing Pearson correlations instead of pairwise accuracies.}
    \label{tab:meta_eval_pearson_results_da_eval_set}
\end{table*}

Table~\ref{tab:meta_eval_results_da_eval_set} compiles the results of the meta-evaluation of a group of reference-based models on the WMT23 DA evaluation set. All of the models are standalone reference-based models. In the table, we contrast four variants of the model fine-tuned in two stages (DA then MQM data) with a model fine-tuned on DA data only (i.e., the first stage only). We present the results on a subset of four language pairs, two of which are present in our MQM training data (en-de and zh-en) and two which are not (en-cs and de-en).

The experiments with mixing DA and MQM data in the second stage of fine-tuning were motivated by the large differences in performance on the WMT23 DA evaluation set observed between a model trained on DA ratings only (row 1 in Table~\ref{tab:meta_eval_results_da_eval_set}) and the same model further fine-tuned on MQM ratings (row 2). As already discussed in~\S\ref{sec:evaluation_data}, this can be partly explained by the discrepancy in DA and MQM rating distributions. This discrepancy understandably affects Pearson correlations, however, it should not have a significant effect on how the metric ranks segments or systems. Nevertheless, while we observed large drops in Pearson's $r$, the pairwise accuracy also dropped substantially for most of the language pairs, both at the segment and the system level. For example, on en-cs the segment-level accuracy drops from $59.54$ to $57.43$, and the system-level accuracy from $87.62$ to $82.86$.

Considering the fact that the performance difference between the models in rows 1 and 2 on en-de and zh-en (i.e., the language pairs with a good amount of MQM training data), are relatively small, we conjecture that further fine-tuning on MQM data alone causes the model to partially ``forget'' other languages from the first stage of fine-tuning. We attempt to prevent the model from this sort of forgetting by mixing some DA ratings into the training set in the second stage.

As the scores of the model in row 3 in the table demonstrate, we are able, for the most part, to restore the performance observed in the first stage of fine-tuning by adding a small proportion of DA training data in the second stage too. Adding not only the DA data, but also the synthetic data, in the second stage (row 5) sometimes boosts the performance further, significantly improving even over the first-stage performance (row 1). Most importantly, the gains over fine-tuning on MQM data alone (row 2) are achieved not at the expense of the model's performance on the MQM or the synthetic test set, as evidenced by the results in Tables~\ref{tab:synthetic_results} and~\ref{tab:meta_eval_results}.

Finally, Table~\ref{tab:meta_eval_pearson_results_da_eval_set} shows the expected big drops in Pearson correlation with the DA ratings after fine-tuning on MQM data (see rows 1 and 2), especially at the segment level. Adding DA data in the second stage helps recover most of the performance (compare rows 3 and 5 with row 1), but as expected, the correlations remain lower particularly for language pairs present in the MQM data the model is fine-tuned on in the second stage (en-de and zh-en).

\subsection{QE Models}
\label{app_sec:qe_results}

In Tables~\ref{tab:synthetic_results_qe} and~\ref{tab:meta_eval_results_qe}, we present the meta-evaluation results for our QE models. These are analogous to those presented in~\S\ref{sec:results}, only the hybrid model is evaluated in a reference-free mode, and the non-hybrid models are ones trained on the source and hypothesis segments only. Note that the hybrid model is the same checkpoint as the one for which we reported the reference-based results in Tables~\ref{tab:synthetic_results} and~\ref{tab:meta_eval_results}, i.e., not one optimized for QE performance.

\begin{table*}
    \small
    \centering
    \begin{tabular}{>{\centering\arraybackslash} m{0.08\linewidth} >{\centering\arraybackslash} m{0.05\linewidth} >{\centering\arraybackslash} m{0.075\linewidth} >{\centering\arraybackslash} m{0.075\linewidth} >{\centering\arraybackslash} m{0.076\linewidth} >{\centering\arraybackslash} m{0.075\linewidth} >{\centering\arraybackslash} m{0.076\linewidth} >{\centering\arraybackslash} m{0.075\linewidth} >{\centering\arraybackslash} m{0.076\linewidth} >{\centering\arraybackslash} m{0.075\linewidth}}
        \toprule
        \textbf{MetricX variant} & \textbf{+DA} & \textbf{+Synth} & \textbf{Empty transl.} & \textbf{Gib-berish} & \textbf{Unre-lated} & \textbf{Under-transl.} & \textbf{Dupli-cation} & \textbf{Missing punct.} & \textbf{Ref-match} \\
        \midrule
        23 & -- & $\sim$ & \textbf{100.00} & \textbf{99.86} & 96.43 & 63.25 & 88.29 & 69.93 & 63.00 \\
        \midrule
        24 & \checkmark & \checkmark & 97.86 & \textbf{99.86} & \textbf{99.43} & \textbf{98.50} & \textbf{98.14} & 65.36 & 63.43 \\
        \midrule
        \multirow{4}{*}{\rotatebox[origin=c]{90}{24-Hybrid}} & -- & -- & 69.86 & \textbf{99.86} & 82.43 & 81.25 & 63.00 & \textbf{77.78} & 63.00 \\
        & \checkmark & -- & 66.14 & 99.57 & 95.29 & 93.50 & 97.86 & 73.86 & 62.57 \\
        & -- & \checkmark & 93.57 & 99.71 & 99.29 & 96.50 & 84.43 & 69.28 & 62.14 \\
        & \checkmark & \checkmark & 93.71 & \textbf{99.86} & \textbf{99.43} & 97.25 & \textbf{98.14} & 69.28 & \textbf{64.14} \\
        \bottomrule
    \end{tabular}
    \vspace{-0.05in}
    \caption{Accuracy of reference-free (QE) MetricX variants in all 7 categories of our synthetic test set. ``23'' is the baseline, the last row of ``24-Hybrid'' corresponds to our primary submission, and ``24'' is our secondary submission. The hybrid model is the same as in Table~\ref{tab:synthetic_results}, only evaluated without references provided as input.}
    \label{tab:synthetic_results_qe}
\end{table*}

\begin{table*}
    \small
    \centering
    \begin{tabular}{>{\centering\arraybackslash} m{0.08\linewidth} >{\centering\arraybackslash} m{0.05\linewidth} >{\centering\arraybackslash} m{0.075\linewidth} >{\centering\arraybackslash} m{0.066\linewidth} >{\centering\arraybackslash} m{0.06\linewidth} >{\centering\arraybackslash} m{0.067\linewidth} >{\centering\arraybackslash} m{0.06\linewidth} >{\centering\arraybackslash} m{0.066\linewidth} >{\centering\arraybackslash} m{0.06\linewidth} >{\centering\arraybackslash} m{0.067\linewidth} >{\centering\arraybackslash} m{0.06\linewidth}}
        \toprule
        \multirow{2}{*}{\textbf{\makecell{MetricX\\variant}}} & \multirow{2}{*}{\textbf{+DA}} & \multirow{2}{*}{\textbf{+Synth}} & \multicolumn{4}{c}{\textbf{Segment-level pairwise accuracy}} & \multicolumn{4}{c}{\textbf{System-level pairwise accuracy}} \\
        & & & \textbf{en-de} & \textbf{zh-en} & \textbf{zh-en\textsuperscript{\textdagger}} & \textbf{en-zh} & \textbf{en-de} & \textbf{zh-en} & \textbf{zh-en\textsuperscript{\textdagger}} & \textbf{en-zh} \\
        \midrule
        23 & -- & $\sim$ & 59.57 & 52.64 & 52.89 & 54.47 & 92.42 & 86.67 & 85.83 & 74.36 \\
        \midrule
        24 & \checkmark & \checkmark & 59.70 & \textbf{54.30} & \textbf{54.48} & 56.00 & 98.48 & \textbf{92.38} & 90.83 & \textbf{87.18} \\
        \midrule
        \multirow{4}{*}{\rotatebox[origin=c]{90}{24-Hybrid}} & -- & -- & 60.11 & 53.80 & 54.00 & \textbf{56.27} & \textbf{100.00} & 89.52 & 89.17 & 84.62 \\
        & \checkmark & -- & 59.18 & 54.08 & 54.30 & 56.14 & \textbf{100.00} & \textbf{92.38} & 90.00 & 84.62 \\
        & -- & \checkmark & \textbf{60.27} & 53.76 & 53.99 & 55.88 & 98.48 & 89.52 & 90.00 & 83.33 \\
        & \checkmark & \checkmark & 59.52 & 54.15 & 54.41 & 55.94 & 98.48 & 90.48 & \textbf{91.67} & 83.33 \\
        \bottomrule
    \end{tabular}
    \vspace{-0.05in}
    \caption{Meta-evaluation scores of reference-free (QE) MetricX variants on the WMT23 MQM evaluation set. ``23'' is the baseline, the last row of ``24-Hybrid'' corresponds to our primary submission, and ``24'' is our secondary submission. The hybrid model is the same as in Table~\ref{tab:meta_eval_results}, only evaluated without references provided as input. \textsuperscript{\textdagger}Alternate references.}
    \label{tab:meta_eval_results_qe}
\end{table*}

Examining first the results on the synthetic test set, summarized in Table~\ref{tab:synthetic_results_qe}, we see similar trends to those observed with reference-based models (Table~\ref{tab:synthetic_results}). The main difference is that the QE models achieve significantly lower performance in the missing punctuation and the reference-matching translation categories. This, however, is expected because both the types of synthetic examples were created from references. In case of the missing punctuation examples, the synthetic translation is simply the reference with the end punctuation removed. Comparing such a hypothesis with the corresponding reference is arguably a significantly easier task than comparing it to the source segment and identifying a missing punctuation symbol. Moreover, there may be a mismatch in the presence of punctuation between the source and the reference in the training examples, making it even more difficult for a QE model to reliably identify missing punctuation. As for the reference-matching translation category, a QE model does not have access to the reference, so it makes perfect sense for it to score a candidate translation better than the reference translation if the reference is of low quality.

Switching over to Table~\ref{tab:meta_eval_results_qe}, which shows the pairwise accuracy of the QE model scores, the trends are also in line with those of the reference-based models in Table~\ref{tab:meta_eval_results}. In contrast to the reference-based results, however, the hybrid model (row 6) does not outperform the standalone model (row 2), although most of the differences are within the expected variance. An astute reader might notice that the accuracy scores on the zh-en test set with the original references and the one with the alternate references do not match (despite the QE models not using the references), and that is because the latter has the original references included as an additional ``human system''.

Finally, we note that our QE models do not fall far behind their reference-based counterparts. In fact, both our primary and secondary QE submissions of MetricX-24 outperform our reference-based MetricX-23 submission from last year, according to the WMT23 MQM evaluation set.

\end{document}